\documentclass{article}

\usepackage[preprint,nonatbib]{neurips_2023}

\usepackage[utf8]{inputenc} 
\usepackage[T1]{fontenc}    
\usepackage{hyperref}       
\usepackage{url}            
\usepackage{booktabs}       
\usepackage{amsfonts}       
\usepackage{nicefrac}       
\usepackage{microtype}      
\usepackage{xcolor}         

\usepackage{amsmath}
\usepackage{amsthm}
\usepackage[pdftex]{graphicx}
\graphicspath{{figures}{../Q (cue)/figures}}
\usepackage{caption}
\usepackage{subcaption}
\usepackage{enumitem}

\newtheorem{claim}{Claim}
\newtheorem{definition}[claim]{Definition}

\newcommand{\X}{\mathcal{X}}
\newcommand{\CB}{\ensuremath{\mathrm{CB}}}

\title{Unsupervised Estimation of Ensemble Accuracy}

\author{%
  Simi Haber \\
  Bar-Ilan University\\
  \texttt{simi@math.biu.ac.il} \\
  \And
  Yonatan Wexler \\
  Q.ai \\
  \texttt{yonatan@q.ai} \\
}

\begin{document}

\maketitle

\begin{abstract}
  Ensemble learning combines several individual models to obtain a better generalization performance.  In this work we present a practical method for estimating the joint power of several classifiers. It differs from existing approaches which focus on ``diversity'' measures by \emph{not relying on labels}.  This makes it both accurate and practical in the modern setting of unsupervised learning with huge datasets.

  The heart of the method is a combinatorial bound on the number of mistakes the ensemble is likely to make. The bound can be efficiently approximated in time linear in the number of samples. We relate the bound to actual misclassifications, hence its usefulness as a predictor of performance.

  We demonstrate the method on popular large-scale face recognition datasets which provide a useful playground for fine-grain classification tasks using noisy data over many classes.
\end{abstract}

\section{Introduction}
Classifier ensembles allow combining several classifiers into a possibly more accurate one.  Traditionally features and ``weak learners'' were used in boosting \cite{AdaBoost, Torralba07, shalevshwartz2011shareboost}.  Nowadays it is common to combine several deep learning networks or random forests into an ensemble classifier \cite{Ensemble06, Rokach10, GANAIE2022105151}.  Each component in the ensemble is chosen according to some specific principle or is randomly selected.  In Ensemble Learning there is empirical evidence that ensembles tend to yield better results when there is significant diversity among the models, with several definitions of the term \cite{Sollich96, Kuncheva03, A:dong_&_yu_&_cao_&_shi_&_ma2020} that are useful when labeled data exists.

In this paper we devise a computationally efficient bound for the number of errors an ensemble will make --- even without using labeled data or joint optimization.  It therefore extends existing approaches to unsupervised learning regime over massive datasets.   This is also true for  many modern datasets which are labeled automatically and contain mistakes such as wrong or duplicate identities \cite{varkarakis2020dataset}.  To the best of our knowledge this is the first time that the idea of ``diversity'' is extended to unsupervised setting.

Intuitively, weak-learners are useful together when they split the input space differently.  Specifically, they need to disagree on their mistakes.  The suggested bound is computed over the observed mapping of the inputs by the classifiers, essentially ignoring their labels.  We show that this information suffices to surface errors the ensemble will produce by posing it as an optimal assignment problem which can be efficiently approximated in $O(N)$ where $N$ is the number of samples.  The bound relies on reasonable assumptions and is verified to work well experimentally.

To give a numeric example, we can check an ensemble of three classifiers over a face recognition dataset with ten million samples of one million identities by picking about a hundred random samples and testing the joint classification on them.

\begin{figure}
  \centering
  \captionsetup{labelfont={bf}}
  \begin{subfigure}[b]{0.245\columnwidth}
    \centering
    \includegraphics[width=\textwidth]{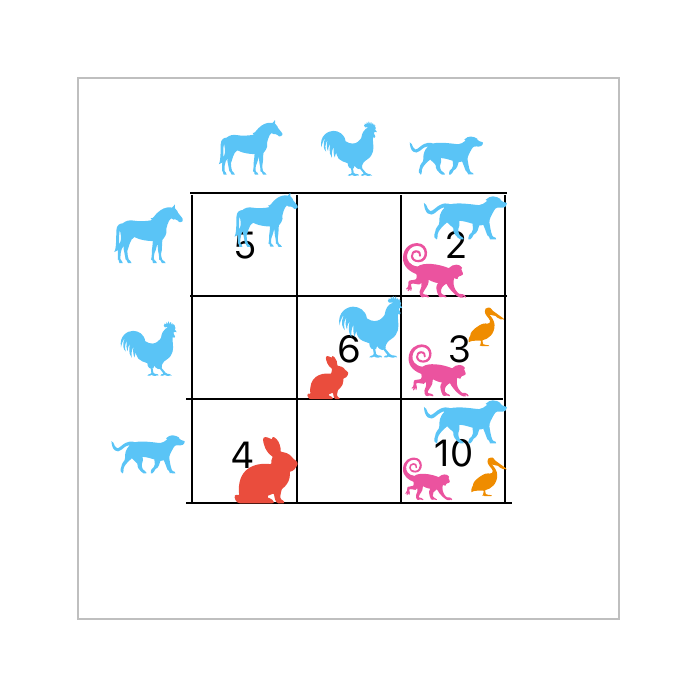}
    \centering
    \captionsetup{justification=centering}
    \caption{Ground truth (unknown)}
    \label{SubFig: ground truth}
  \end{subfigure}
  \hfill
  \begin{subfigure}[b]{0.245\columnwidth}
    \captionsetup{justification=centering}
    \centering
    \includegraphics[width=\textwidth]{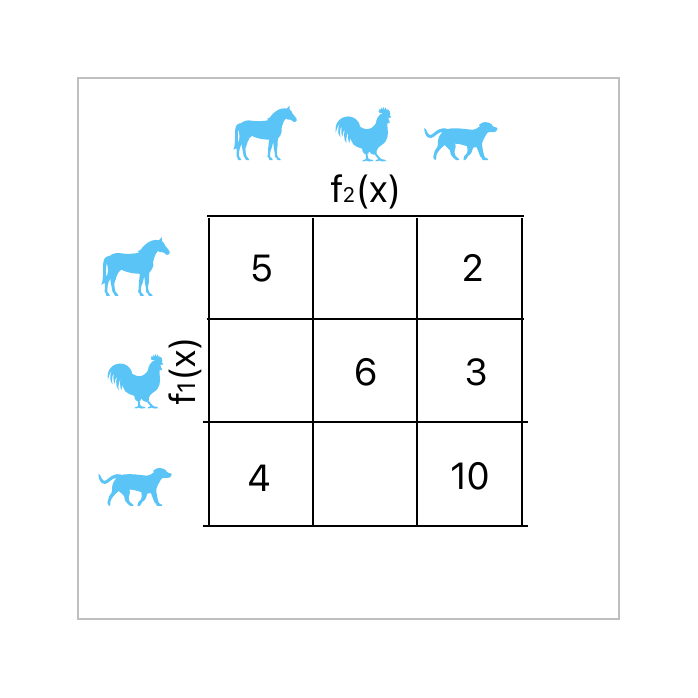}
    \caption{Cell Occupancy $\mathbf{C}$ \\(visible)}
    \label{SubFig: C matrix}
  \end{subfigure}
  \hfill
  \begin{subfigure}[b]{0.245\columnwidth}
    \captionsetup{justification=centering}
    \centering
    \includegraphics[width=\textwidth]{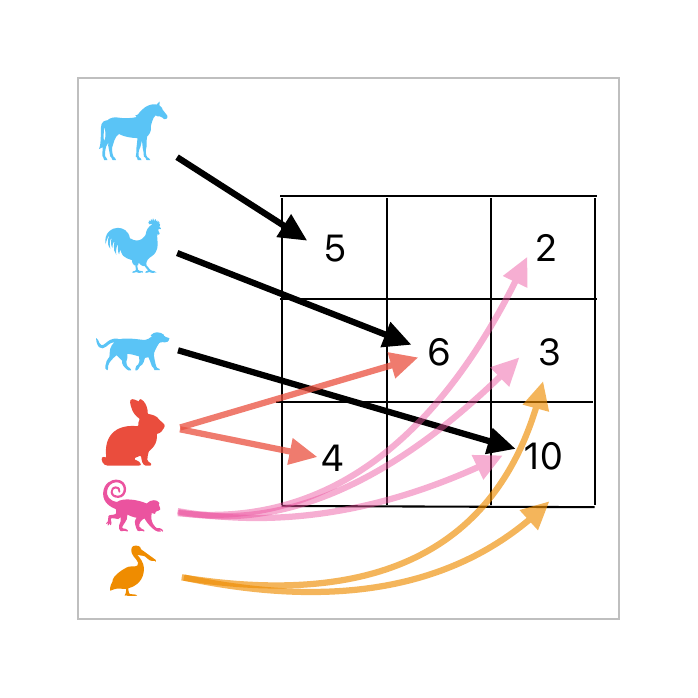}
    \captionsetup{justification=centering}
    \caption{Bipartite graph\\ representation}
    \label{SubFig: bipartite rep}
  \end{subfigure}
  \hfill
  \begin{subfigure}[b]{0.245\columnwidth}
    \captionsetup{justification=centering}
    \centering
    \includegraphics[width=\textwidth]{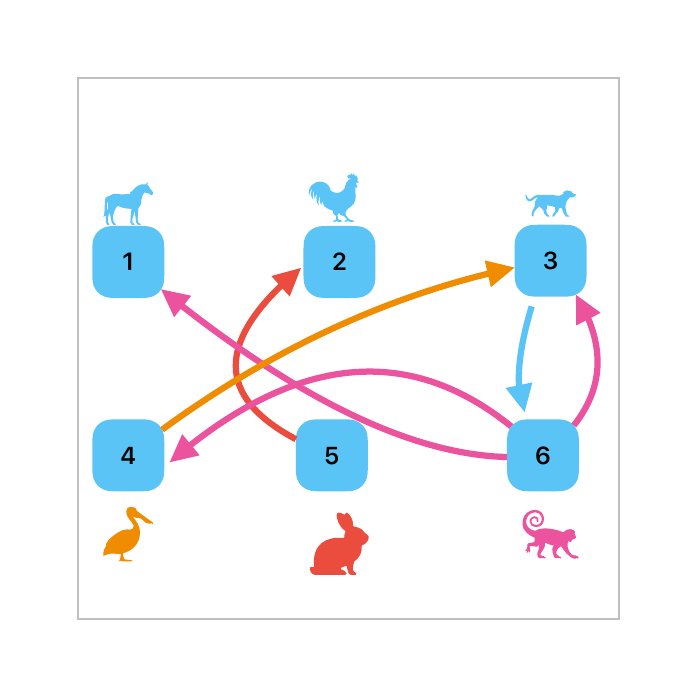}
    \caption{Mistakes\\ multigraph}
    \label{SubFig: mistakes graph}
  \end{subfigure}
  \caption{ A schematic example of the approach where we wish to estimate the potential ensemble performance of $Q=2$ classifiers over $K=6$ classes projected onto $L=3$ classes.
    Panel \textbf{(\subref{SubFig: ground truth})} shows the (unknown) true mapping. Note that monkey class is split among several cells thus limiting the potential accuracy.  Panel \textbf{(\subref{SubFig: C matrix})} shows the count of the number of elements in the mapping $x_i \mapsto \left(f_1(x_i), f_2(x_i) \right)$ which is our only observable.  Panel \textbf{(\subref{SubFig: bipartite rep})} illustrates the bipartite graph of the same mapping where multiplicity of edges is counted.  Panel \textbf{(\subref{SubFig: mistakes graph})} shows the multigraph used in the proof of Claim \ref{Claim: CB monotone with real}.  }
  \label{Fig: detailed example}
\end{figure}

Figure \textbf{\ref{Fig: detailed example}} illustrates the method for an ensemble of two classifiers
$f_1, f_2: \X \mapsto [L]$.  We count duplicity of tuples $ \displaystyle{(f_1(x_i), f_2(x_i))}_{i=1}^N$ creating the matrix $C$ (Panel \textbf{\ref{SubFig: C matrix}}, where rows are the output of $f_1$ and columns of $f_2$).  The count enables efficient approximation of a lower bound for the number of mistakes over all possible mappings from a class into (one or more) matrix cell (Panel \textbf{\ref{SubFig: bipartite rep}}) such that all samples are accounted for.  This without knowing the ground truth (Panel \textbf{\ref{SubFig: ground truth}}).

The paper is organized as follows:  In Section \ref{sec:formulation}, we show how two or more metric-learner based classifiers can be used to construct a joint space partitioning scheme. In Section \ref{Sec: Approach} we formally define and analyze the proposed bound.  In Section \ref{Sec: Experiments} we verify the method on real data and conclude.

\section{Problem Formulation}  \label{sec:formulation}

Let $\{x_i \in \X\}_{i=1}^N$ be samples from $K$ classes. Let $\{f_q\}_{q=1}^Q$ be multi-class classifiers.  We aim to estimate their potential accuracy over the samples without using labels.

Every multi-class classifier partitions $\X$ into (not necessarily connected) parts, one for each class.  Two classifiers splitting the space differently have the potential to jointly refine the partition of $\X$, thus possibly enabling higher accuracy.  At high dimension the number of parts in the joint partition is likely to grow exponentially while the likelihood of a sample to appear in a part diminishes.  We therefore pick a small random set of $L \ll K$ samples from distinct classes and observe the joint behavior of $f_q$ on them.  A good choice is $L > \sqrt[Q] K$ yielding more cells than classes. This forces some cells to be empty in a perfect classification which increases the probability of visible errors.

We start with the following claim:
\begin{claim} \label{claim:mc2ml}
  Every multi-class classifier can be converted to a metric-learner and vice-versa.
\end{claim}
\vskip -3ex \vskip 0pt
\begin{proof}[Proof]
By construction for each direction separately:
\vskip -2ex \vskip 0pt
\begin{description}[leftmargin=0pt]
\item[$\Leftarrow$] Given a multi-class classifier $f(x) \mapsto [K]$ define a metric learner $g(x) := e_{f(x)}$ which maps the class $k$ into the unit vector $e_k\in R^K$ in the standard basis.
\item[$\Rightarrow$]  Given a metric learner $f(x) \mapsto R^d$ satisfying $||f(x)-f(y)|| \leq ||f(x)-f(z)||$ for $x,y$ in one class and $z$ in another, and given a one or more (sample, class) pair from each class $(x_i,y_i)$. Define $g(x) := y_j$ where $j:=\mathrm{argmin}_i ||f(x)-f(x_i)||$.
\end{description}
\vskip -1ex \vskip 0pt
It is immediate to verify that this mapping preserves mistakes and hence also accuracy.
\end{proof}

By claim \ref{claim:mc2ml} we can assume w.l.o.g.\ that $f_q$ are metric-learners.  Therefore any choice of a random set of $L$ samples induces a mapping of samples $x_i \mapsto \vec i\in [L]^Q$ where $\vec i:=\left(\,f_1(x_i), \cdots, f_Q(x_i)\,\right)$.  The structure $\textbf{C}=[c_{\vec i}]$ counts how many samples are mapped into each tuple (also called Cell).  When $Q=2$ the structure $\textbf{C}$ is a matrix. This count is our only observable. We next show it is indicative of ensemble accuracy.

\section{Approach and analysis}\label{Sec: Approach}
The core idea for our error analysis is that any cell $c_{\vec i}$ which is not of a class size must contain an error.  We seek a mapping matching classes to cells such that all samples are accounted for.  Given such mapping we count the number of samples that are mapped incorrectly.  This approach may miss cells that unknowingly mix several classes, a point which we address in \ref{Claim: CB monotone with real}.

We define the bound by considering bipartite multi-graphs (see Panel \textbf{\ref{SubFig: bipartite rep}}) with two types of nodes $N=[K] \cup \{\vec i\}$. On the left the nodes represent the $K$ classes and on the right the nodes represent the $L^Q$ cells. Denote by $H_{\vec i}^k$ the (unknown) number of elements from class $k$ that are mapped into cell $\vec i$.   Out of all possible bipartite multigraphs on this vertex set with left degrees $S$ and right degrees $c_{\vec i}$, we seek $H^*$ maximizing $\sum_{\vec i}\sum_k (H_{\vec i}^k)^2$.  The objective function counts pairs of samples of the same class that end up in the same cell and penalizes the rest.

This leads to the following definition:
\begin{definition}
Given a list of cell sizes $\mathbf{C}$ and class size $S$ we define
\[
\CB(\mathbf{C}) = \max_H \sum_{\vec i}\sum_k (H_{\vec i}^k)^2 ,
\quad \text{s.t.}\quad
\sum_{\vec i} H_{\vec i}^k = S \,,\, \sum_k H_{\vec i}^k = c_{\vec i} .
\]

\end{definition}
Computing $\CB(\mathbf{C})$ is NP-hard, being a generalization of the Multiway Number Partitioning problem \cite{A:graham1969}. Yet it is sometimes called ``The easiest NP-hard problem'' \cite{IC:mertens2006} as it can be approximated efficiently. In some cases we can find $\CB(\mathbf{C})$ exactly:

\begin{claim}
There is a pseudopolynomial algorithm for exact computation of $\CB(\mathbf{C})$ requiring $O(L^2(K_r-1)C_m^{K_r-1})$ memory where $C_m$ is the size of the largest cell and $K_r$ is the number of classes after removal of classes and cells of equal size.
\end{claim}
The proof follows \cite{IP:korf2009} with two adaptations. First, as long as there is a cell of size $S$, we remove it from $\textbf{C}$. The second modification is that we allow cells to be split between classes and vice-versa. As $K$ tends to be large the above does not provide a practical solution in most cases.

\begin{claim}
There is a polynomial time approximation scheme (PTAS) for finding $\CB(\mathbf{C})$ with runtime $O(K_r)$ with exponential dependence in the reciprocal of the precision.
\end{claim}
Here we apply a PTAS of Alon et al. \cite{A:alon_&_azar_&_woeginger_&_yadid1998}, again after removal of cells and classes of equal size.

Finally, a greedy algorithm gives a $4/3$ approximation ratio \cite{A:graham1969}. Our experiments demonstrate this suffices in some cases.

We next prove that the number of ``hidden'' mistakes is likely to grow as $\CB(\mathbf{C})$ grows, something that is also shown empirically in Section \ref{Sec: Experiments}.
Intuitively, some misclassifications may change cell counts and be visible to us, while others are not.  We do not expect classifiers to have a tendency towards detectable vs.\ invisible errors, so it is not surprising the two grow together. Hence we hypothesize that the number of visible errors increases monotonically with the total number of mistakes.

Let $\varphi^*$ be an optimal mapping of classes to cells achieving $\CB(\mathbf{C})$.  Whenever a sample from class $c$ is mapped to a different cell than $\varphi^*(c)$ we consider it a mistake. We can show the following:
\begin{claim}\label{Claim: CB monotone with real}
Let $F$ be a classifier and assume that when $F$ misclassifies, the target class is chosen uniformly at random. Then the expectation of $\CB(\mathbf{C})$ is monotonically increasing with the total number of mistakes.
\end{claim}
For the proof we define the mistakes graph --- a directed multigraph where each sample not mapped into the correct cell is represented by an arc between the correct cell and the mapped cell (see Fig.\ \textbf{\ref{SubFig: mistakes graph}}). Hidden mistakes are those appearing in a cycle or in a path besides the first and last edges. Hence counting such errors amounts to finding an edge maximum cover of the multi-graph by edge disjoint directed cycles and paths. We complete the proof by analyzing edge-maximal edge-disjoint collection of cycles in the appropriate random graph model

The approach of the last proof yields a polynomial relation: $\CB(\mathbf{C}) = O(f^c)$ where $f$ is the number of errors and $c>1/5$ is constant. We believe that a stronger result, $c\geq 1/2$, may be obtained by finding an analog to the main result of \cite{A:mckay_&_wormald1997} for the mistakes graph --- showing essential independence between degrees and analyzing cell size distribution widths. In silico, when using two similar classifiers, we get heavy counts on the diagonal of $\mathbf{C}$, in which case the number of visible errors grows linearly with the number of total mistakes.
\vskip -2ex \vskip 0pt
\section{Experiments}\label{Sec: Experiments}
\vskip -1ex \vskip 0pt
\begin{figure}[t]
  \centering
  \captionsetup{labelfont={bf}}
  \begin{subfigure}[b]{0.49\columnwidth}
    \centering
    \includegraphics[width=\textwidth]{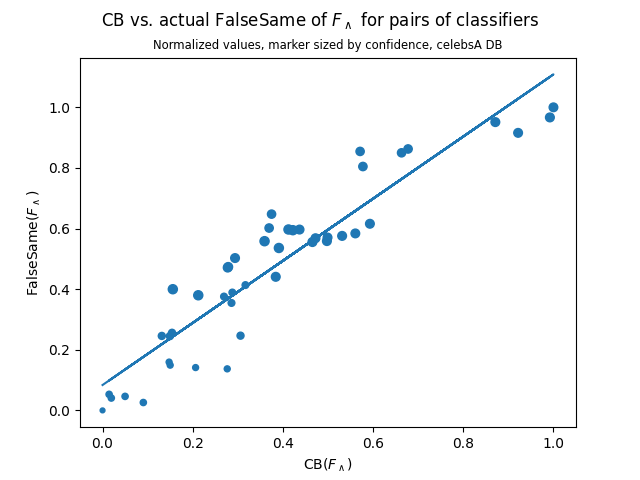}
    \caption{CelebsA dataset}
    \label{SubFig: real vs CB, celebA}
  \end{subfigure}
  \hfill
  \begin{subfigure}[b]{0.49\columnwidth}
    \centering
    \includegraphics[width=\textwidth]{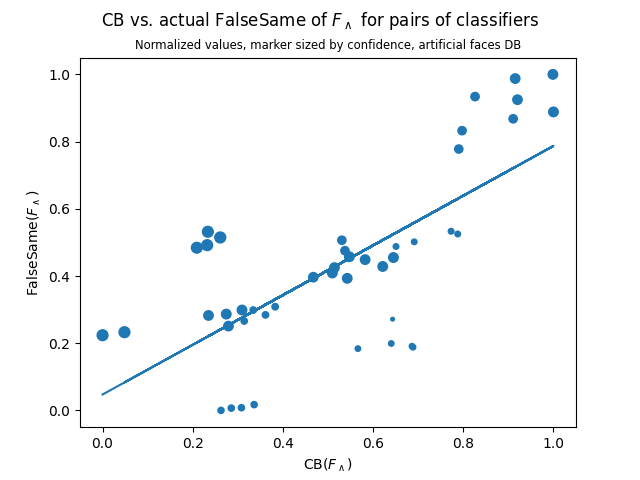}
    \caption{artificial faces dataset - DigiFace1M}
    \label{SubFig: real vs CB, DigiFace}
  \end{subfigure}
  \caption{Real True-same counts vs.\ $\CB$ estimation}
  \label{Fig: real vs CB}
\end{figure}
We created a collection of ten classifiers $f_q$ and measured $\CB$ for all $45$ pairs on two datasets. We have used the \texttt{Dlib} \cite{dlib09, dlib} as a base. This metric learner maps a photo into ${\mathcal R}^{128}$. The authors recommend a distance threshold of $0.6$ when deciding if two photos are of the same origin. Representatives from $L$ random classes were randomly chosen, and proximity to them was used in classification. We have created our metric learners in a few different ways:
\vskip -2ex \vskip 0pt
\begin{itemize}[leftmargin=*]
\item $d$ pairs of samples were randomly chosen. For each pair we took the line connecting the two samples and projected each sample on it, with an affine correction making the center of the segment the origin. This projection bisects the samples along probable interesting affine hyperplanes.
\vskip -1ex \vskip 0pt
\item Using only randomly chosen $d$ coordinated from the original 128.
\vskip -1ex \vskip 0pt
\item Same as above, but using more representatives from each class.
\end{itemize}
\vskip -2ex \vskip 0pt
Table \textbf{\ref{table: datasets}} lists the datasets we used. Figure \textbf{\ref{Fig: real vs CB}} presents the results, Panel \textbf{\ref{SubFig: real vs CB, celebA}} for the CelebsA dataset \cite{liu2015faceattributes} and Panel \textbf{\ref{SubFig: real vs CB, DigiFace}} for DigiFace1M \cite{bae2023digiface1m}. In both charts the $x$-axis is $\CB$ and $y$-axis is the real False-Same error measured using labels. Pairs of samples were considered Same when both classifiers agreed they are the same. Markers are sized according to the sum of real performances of each classifier. Regression lines were added for convenience.
\begin{table}[th]
  \small
  \captionsetup{labelfont={bf}}
  \caption{Datasets characteristics}
  \label{table: datasets}
  \centering
  \begin{tabular}{lrrll}
    \toprule
    Name       & Identities & Samples    & Reference & Remarks \\
    \midrule
    CelebaA    &   10,000   &    200,000 & \cite{liu2015faceattributes} &  \\
    DigiFace1M   &  100,000   &  1,220,000 & \cite{bae2023digiface1m} & Artificially generated  \\
    \bottomrule
  \end{tabular}
\end{table}
\vskip -3ex \vskip 0pt
\section{Conclusion}\label{Sec: Conclusion}
We presented a combinatorial bound on the number of mistakes an ensemble is likely to make, without relying on labeled data.  The bound can be approximated efficiently while still agreeing with real data.  As such it is useful in modern settings of huge datasets in the unsupervised learning regime.

Due to space limitations we left out some proofs to be added in a longer version of this manuscript. We would like to thank the anonymous referee for keen reading and insightful suggestions that improved this manuscript.


\begin{thebibliography}{10}

\bibitem{A:alon_&_azar_&_woeginger_&_yadid1998}
Noga Alon, Yossi Azar, Gerhard~J. Woeginger, and Tal Yadid.
\newblock Approximation schemes for scheduling on parallel machines.
\newblock {\em Journal of Scheduling}, 1:55--66, December 1998.

\bibitem{bae2023digiface1m}
Gwangbin Bae, Martin de~La~Gorce, Tadas Baltru{\v{s}}aitis, Charlie Hewitt,
  Dong Chen, Julien Valentin, Roberto Cipolla, and Jingjing Shen.
\newblock Digiface-1m: 1 million digital face images for face recognition.
\newblock In {\em 2023 IEEE Winter Conference on Applications of Computer
  Vision (WACV)}. IEEE, 2023.

\bibitem{A:dong_&_yu_&_cao_&_shi_&_ma2020}
Xibin Dong, Zhiwen Yu, Wenming Cao, Yifan Shi, and Qianli Ma.
\newblock A survey on ensemble learning.
\newblock {\em Frontiers of Computer Science}, 14(2):241--258, April 2020.

\bibitem{AdaBoost}
Yoav Freund and Robert~E. Schapire.
\newblock A desicion-theoretic generalization of on-line learning and an
  application to boosting.
\newblock In Paul Vit{\'a}nyi, editor, {\em Computational Learning Theory},
  pages 23--37, 1995.

\bibitem{GANAIE2022105151}
M.A. Ganaie, Minghui Hu, A.K. Malik, M.~Tanveer, and P.N. Suganthan.
\newblock Ensemble deep learning: A review.
\newblock {\em Engineering Applications of Artificial Intelligence},
  115:105151, 2022.

\bibitem{A:graham1969}
Ron~L. Graham.
\newblock Bounds on multiprocessing timing anomalies.
\newblock {\em SIAM Journal on Applied Mathematics}, 2(17):416--429, March
  1969.

\bibitem{dlib}
Davis~E. King.
\newblock Dlib c++ library.
\newblock \url{http://dlib.net}.
\newblock Version 19.24; Accessed: 2023-08-31.

\bibitem{dlib09}
Davis~E. King.
\newblock Dlib-ml: A machine learning toolkit.
\newblock {\em Journal of Machine Learning Research}, 10:1755--1758, 2009.

\bibitem{IP:korf2009}
Richard~E. Korf.
\newblock Multi-way number partitioning.
\newblock In {\em IJCAI-09}, pages 538--543, 2009.

\bibitem{Kuncheva03}
Ludmila~I. Kuncheva and Christopher~J. Whitaker.
\newblock Measures of diversity in classifier ensembles and their relationship
  with the ensemble accuracy.
\newblock {\em Machine Learning}, 51:181--–207, 2003.

\bibitem{liu2015faceattributes}
Ziwei Liu, Ping Luo, Xiaogang Wang, and Xiaoou Tang.
\newblock Deep learning face attributes in the wild.
\newblock In {\em Proceedings of International Conference on Computer Vision
  (ICCV)}, December 2015.

\bibitem{A:mckay_&_wormald1997}
Brendan~D. McKay and Nicholas~C. Wormald.
\newblock The degree sequence of a random graph. {I}. {T}he models.
\newblock {\em Random Structures \& Algorithms}, 11(2):97--117, September 1997.

\bibitem{IC:mertens2006}
Stephan Mertens.
\newblock The easiest hard problem: Number partitioning.
\newblock In Allon~G. Percus, Gabriel Istrate, and Cristopher Moore, editors,
  {\em Computational Complexity and Statistical Physics}, pages 125--139.
  Oxford University Press, 2006.

\bibitem{Ensemble06}
Robi Polikar.
\newblock Ensemble based systems in decision making.
\newblock {\em IEEE Circuits and Systems Magazine}, 6(3):21--45, 2006.

\bibitem{Rokach10}
Lior Rokach.
\newblock Ensemble-based classifiers.
\newblock {\em Artif. Intell. Rev.}, 33:1--39, 02 2010.

\bibitem{shalevshwartz2011shareboost}
Shai Shalev-Shwartz, Yonatan Wexler, and Amnon Shashua.
\newblock Shareboost: Efficient multiclass learning with feature sharing, 2011.

\bibitem{Sollich96}
Peter Sollich and Anders Krogh.
\newblock Learning with ensembles: How overfitting can be useful.
\newblock {\em Advances in Neural Information Processing Systems}, 8:190--196,
  1996.

\bibitem{Torralba07}
Antonio Torralba, Kevin~P. Murphy, and William~T. Freeman.
\newblock Sharing visual features for multiclass and multiview object
  detection.
\newblock {\em IEEE Transactions on Pattern Analysis and Machine Intelligence},
  29(5):854--869, 2007.

\bibitem{varkarakis2020dataset}
Viktor Varkarakis and Peter Corcoran.
\newblock Dataset cleaning -- a cross validation methodology for large facial
  datasets using face recognition, 2020.

\end{thebibliography}
\end{document}